\documentclass[conference]{IEEEtran}
\IEEEoverridecommandlockouts
\usepackage{cite}
\usepackage{amsmath,amssymb,amsfonts}
\usepackage{algorithmic}
\usepackage{graphicx}
\usepackage{textcomp}
\usepackage{xcolor}
\usepackage{multirow}
\usepackage{array}

\def\BibTeX{{\rm B\kern-.05em{\sc i\kern-.025em b}\kern-.08em
    T\kern-.1667em\lower.7ex\hbox{E}\kern-.125emX}}
\begin{document}

\title{Global Spatio-Temporal Fusion-based Traffic Prediction Algorithm with Anomaly Aware	\thanks{This research was supported by the National Natural Science Foundation of China under Grant 61906038, Grant 62336003 and Grant 62006119 (Corresponding author: Guangyu Li). Chaoqun Liu, Chen Gong, and Guangyu Li are with PCA Lab, Key Lab of Intelligent Perception and Systems for High-Dimensional Information of Ministry of Education, and Jiangsu Key Lab of Image and Video Understanding for Social Security, Nanjing University of Science and Technology, and Xuanpeng Li is from Southeast University, Nanjing 210094, China .(e-mail: Liuchaoqun@njust.edu.cn; li\_xuanpeng@seu.edu.cn; guangyu.li2017@njust.edu.cn; chen.gong@njust.edu.cn).}
}
\author{\IEEEauthorblockN{Chaoqun Liu, Xuanpeng Li, Chen Gong, Guangyu Li*}
}

\maketitle

\begin{abstract}
Traffic prediction is an indispensable component of urban planning and traffic management. Achieving accurate traffic prediction hinges on the ability to capture the potential spatio-temporal relationships among road sensors. However, the majority of existing works focus on local short-term spatio-temporal correlations, failing to fully consider the interactions of different sensors in the long-term state. In addition, these works do not analyze the influences of anomalous factors, or have insufficient ability to extract personalized features of anomalous factors, which make them ineffectively capture their spatio-temporal influences on traffic prediction. To address the aforementioned issues, We propose a global spatio-temporal fusion-based traffic prediction algorithm that incorporates anomaly awareness. Initially, based on the designed anomaly detection network, we construct an efficient  anomalous factors impacting module (AFIM), to evaluate the spatio-temporal impact of unexpected external events on traffic prediction. Furthermore, we propose a multi-scale spatio-temporal feature fusion module (MTSFFL) based on the transformer architecture, to obtain all possible both long and short term correlations among different sensors in a wide-area traffic environment for accurate prediction of traffic flow. Finally, experiments are implemented based on real-scenario public transportation datasets (PEMS04 and PEMS08) to demonstrate that our approach can achieve state-of-the-art performance.
\end{abstract}


\section{Introduction}
\begin{figure}[ht]
	\centering
	\includegraphics[trim=110 150 110 150, clip, width=0.9\columnwidth]{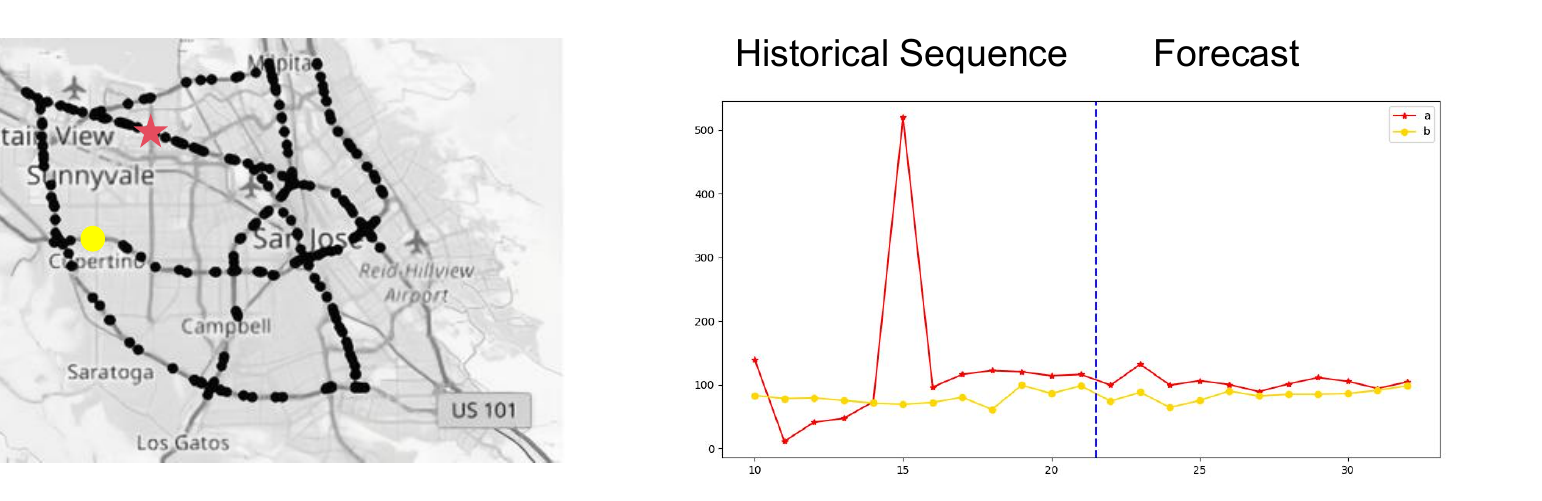}  
	\caption{Scenario Description.}
	\label{Description}
\end{figure}

Accurate traffic flow prediction can effectively reduce traffic congestion and alleviate traffic pressure \cite{Yigit2023}, which is crucial for the development of smart transportation systems \cite{GWNet}. As shown in Fig \ref{Description}, traffic prediction is a typical spatio-temporal sequence prediction problem that involves analyzing complex spatio-temporal correlations. Additionally, traffic prediction is influenced by various factors \cite{Yigit2022}, which makes accurate forecasting challenging \cite{DCRNN}. Consequently, researchers have extensively studied methods to improve the accuracy of traffic state predictions \cite{STID}. Some researchers have employed statistical modeling methods for traffic predictions \cite{GMAN}. However, these approaches only consider historical data from individual road sensors and overlook complex inter-sensor correlations, leading to poor prediction outcomes \cite{PDFormer}.
In recent years, deep learning-based methods have become increasingly popular in the field of traffic flow prediction \cite{MTGNN}. Wang et al. \cite{TGAE} demonstrated the efficacy of encoding road structure information into the latent space using directed-attention neighborhood aggregation and highlighted the potential of combining this approach with Long Short-Term Memory (LSTM) networks to model the temporal evolution of traffic. The Long Short-Term Memory (LSTM) network effectively preserves the nonlinear temporal dependence of node embeddings across successive time steps, leading to improved performance in traffic prediction tasks. Liu et al. \cite{STAE} proposed a novel spatio-temporal adaptive embedding and integrated it with a periodic embedding into a transformer-based model, achieving excellent results in traffic prediction. Li et al. \cite{STFGNN} extended the graph structure into a spatio-temporal graph by combining data from neighboring time steps, and then extracted both long and short term correlations among traffic nodes using graph convolution on this spatio-temporal graph, which improved the accuracy of traffic prediction. However, these methods primarily concentrate on the transient correlations between neighboring or local spatial nodes, and are therefore inadequate for the extraction of complex spatio-temporal correlations in the context of large-scale road environments, which would result in a significant decline in model performance with increasing in prediction time.
\begin{figure*}[ht]
	\centering
	\includegraphics[width=2.0\columnwidth, trim=0.5cm 0.5cm 0.5cm 0.5cm, clip]{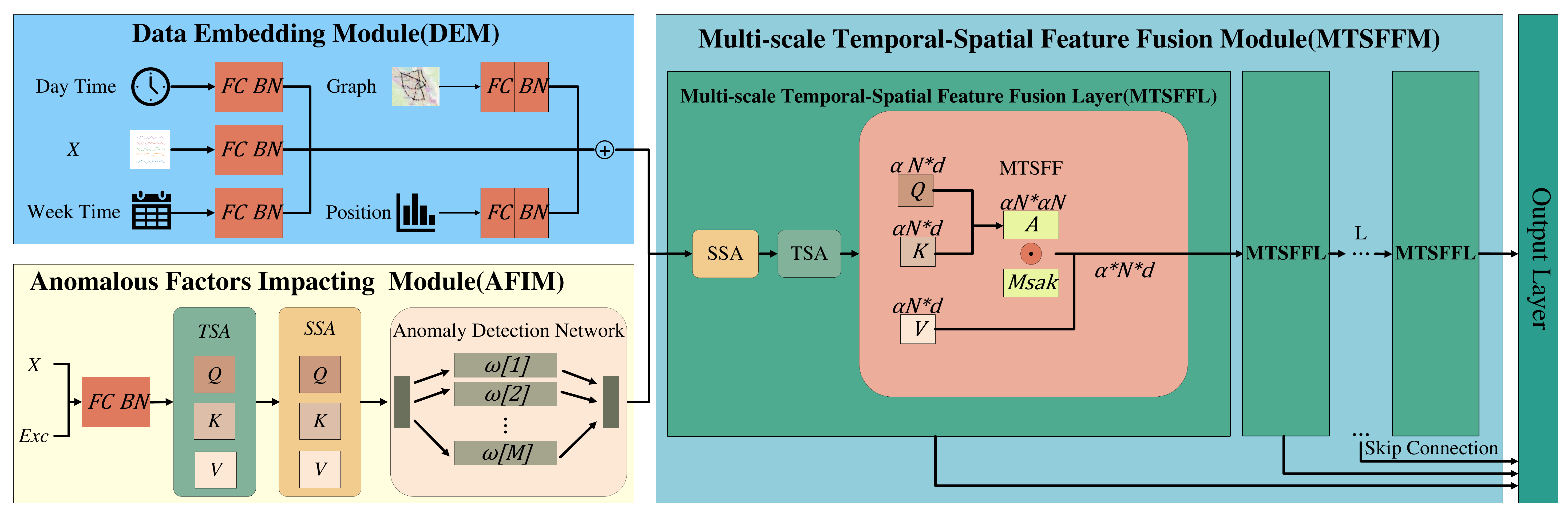}
	\caption{The overall framework of GSTF.}
	\label{AASTMA}
\end{figure*}

Additionally, the traffic environment is affected by numerous external factors, such as weather, traffic accidents, and road construction, leading to irregular fluctuations in traffic patterns and complicating traffic prediction \cite{Ozcevik2017}. To address this problem, Essien et al. proposed a traffic forecasting algorithm based on a Stacked Autoencoder (SAE) architecture with Long Short-Term Memory (LSTM) to combine information extracted from Twitter messages with weather data \cite{URBAN}. Zhu et al. \cite{KST-GCN} represented traffic information and various external factors as a heterogeneous semantic network, and then employed knowledge graph representation methods to capture the knowledge structures and semantic relationships between traffic information and external factors to achieve accurate traffic prediction. Han et al. \cite{UAGCRN} improved prediction accuracy by leveraging human activity frequency data from a household travel survey to capture the relationships between traffic patterns and human activities. However, the existing works treat the anomalous factors in the same way as the original data, and the generalized feature extraction networks are not capable of effectively extracting the special spatio-temporal influences of the anomalous factors on the traffic prediction, which in turn leads to a poor performance of the traffic prediction.

In order to effectively capture the global long-time information as well as the personalized characteristics of anomalous factors, we propose a global spatial-temporal fusion traffic prediction algorithm with anomaly awareness. Firstly, based on the anomaly classification network, we design a feasible  anomaly influence module, and input the both traffic state information and external anomaly information to obtain the anomaly influence characterization. Secondly, we integrated temporal, location, and spatial embeddings to form a high-dimensional representation and output, so as to obtain richer feature representations. Furthermore, we construct an attention-based multiscale spatio-temporal feature mixing module, to obtain the long/short time feature information of the global traffic state and achieve accurate traffic prediction results. The principal contributions of this study can be summarized in three points.
\begin{itemize}
\item Based on an anomaly detection network, we construct an anomalous factors impacting  module to accurately quantify its complex spatio-temporal impacts on traffic prediction.
\item By means of the proposed spatio-temporal self-attention mechanism architecture, we design multi-scale spatio-temporal feature fusion module to explicitly capture both short and long spatio-temporal relationships among different sensors across a variety of scenarios. 
\item Quantitative and qualitative experiments are conducted on two benchmark datasets (PEMS04 and PEMS08), demonstrating that our approach  outperforms all state-of-the-art comparative methods.
\end{itemize}
\section{PRELIMINARIES}
Traffic prediction, as a type of spatio-temporal prediction, is the process of forecasting the future state of traffic based on historical data. This data includes the impact of external anomalous events, such as traffic accidents, road construction, and other factors. A traffic network is defined as a graph $ \mathbf{G} = (\mathbf{N}, \mathbf{E}, \mathbf{A}) $
, where $ \mathbf{N} $ represents the set of sensors on a traffic road, $ \mathbf{E} $ represents the connectivity between traffic sensors (i.e., roads), and $ \mathbf{A} $ represents the adjacency matrix, which is $ 1 $ when two sensors are connected by a road and $ 0 $ otherwise. $ N=|\mathbf{N}| $ represents the number of sensors. Furthermore, we define the traffic state information on the road as a tensor $ X \in \mathbb{R}^{T \times N \times C} $
, where $ T $ represents the time interval and $ C $ represents the number of traffic states (such as traffic flow, traffic speed). Additionally, we define the external abnormal event impact data as a tensor $ Exc \in \mathbb{R}^{T \times N \times C} $. In conclusion, the traffic prediction problem can be modeled as
\begin{equation}
	F([X_{t-\alpha+1},...,X_t,\mathbf{G},Ext])= [X_{t+1},...,X_{t+\beta}]  
\end{equation}
where $F(.)$ denotes the designed deep learning model, $ \alpha $ is the duration for which the historical data sequence, $ \beta $ indicates the  prediction period, $X_t$ refers to the traffic state from all traffic sensors at a specific point in time.
\section{METHODOLOGY}
To fully explore the multi-scale spatiotemporal influences between traffic state data (including global/local traffic scenes, long/short time lengths) and capture the impact of external anomalous factors on road nodes, we propose a traffic prediction algorithm based on global spatiotemporal fusion. Our method mainly consists of the following parts: the first part is the Data Embedding Module(DEM), the second part is the Anomalous Factors Impacting  Module, and the third part is the Multi-scale Spatiotemporal Feature Fusion Module, and the main framework is illustrated in Fig \ref{AASTMA}.
\subsection{Anomalous Factor Impact Module}
In real-world road scenarios, various external anomalous factors, such as traffic accidents or on-site construction, can have a significant spatiotemporal impact on traffic conditions. These events can not only affect the state of current road sensors but also, over time, extend their influence to distant connected nodes. To effectively model the impact of anomalous factors, we propose an Anomalous Factor Impact Module based on an anomaly classification network. This module includes three components: Temporal Self-Attention (TSA), Spatial Self-Attention (SSA), and Anomaly Detection Network.

(1) Embedding Representation of Anomalous Information

Due to the paucity of datasets containing multiple types of external anomalous events, we utilize a moving average method to identify anomalous data from raw data and label these points as anomalies. If a data point is deemed anomalous, it is marked as 1; otherwise, it is marked as 0. Thus, we obtain labeled data for the anomaly factors, consistent with the shape of the input $X$, denoted as $Exc \in \mathbb{R}^{\alpha \times N \times C}$. Furthermore, in order to effectively capture the impact of anomaly factors from the original information, we concatenate $Exc$ with the original data and map it to a high-dimensional space to obtain an embedded representation of the anomaly information, denoted as $Exc_{emb} \in \mathbb{R}^{\alpha \times N \times d}$. The specific formula is as follows:
\begin{equation}
	Exc\_emb=BN(FC(cat[X,Exc]))
\end{equation}
where $X$ stands for the input historical traffic state data, $Exc$ denotes the labeled anomalous factor data, $FC$ signifies the fully connected layer, and $BN$ refers to the batch normalization layer.

(2) Temporal Impact of External Anomalous Factors

To extract the temporal influence of external anomaly factors, we designed a Temporal Self-Attention component (TSA). First, we reshape the obtained $Exc_{emb}$, changing its tensor shape $\alpha \times N \times d$ to $N \times \alpha \times C$, to calculate the influence of anomaly factors along the temporal dimension. Second, using a fully connected layer, we generate the query and key for each node $i$ at each time step, allowing us to compute the similarity between different time points, resulting in a similarity matrix $A_{T}\in \mathbb{R}^{\alpha \times \alpha}$, which can be expressed as 
\begin{equation}
	A_{T}=\frac{Q_i^T{{(K}_i^T)}^\mathcal{T}}{\sqrt{\tilde{d}}}
\end{equation}
where $ Q_i^T=Exc\_emb_{i::}W_q^T \in \mathbb{R}^{\alpha \times \tilde{d}}$ represents the time query matrice obtained through the fully connected layer, $K_i^T=Exc\_emb_{i::}W_k^T \in \mathbb{R}^{\alpha \times \tilde{d}}$ indicates the time key matrice obtained through the fully connected laye, and $\tilde{d}$ stands for the hidden dimension in the self-attention mechanism.  

Finally, by multiplying the similarity matrix with the value matrix obtained through the fully connected layer, we derive the temporal influence of external anomaly factors. The specific representation is as follows:
\begin{equation}
	\text{TSA}({Exc\_emb}_{i::})=FC(softmax(A_{T})V_i^T\ )
\end{equation}
where $V_i^T=Exc\_emb_{i::}W_v^T \in \mathbb{R}^{\alpha \times d}$ represents the time value matrix obtained from the fully connected layer, $ FC $ refers to the fully connected layer. By applying the above operations to all sensor data, the output of the TSA component is obtained$ Exc_T\in \mathbb{R}^{N \times \alpha \times d} $.
\begin{equation}
	Exc\_T=\text{TSA}(Exc\_emb)
\end{equation}
(3) Spatial Impact of External Anomalous Factors

The influence of external anomaly factors spreads over time to other connected road sensors. Therefore, we design a Spatial Self-Attention component (SSA) based on road links, incorporating a masked mechanism. First, we reshape the obtained tensor $ Exc_T\in \mathbb{R}^{N \times \alpha \times d} $ , transforming its shape from $N \times \alpha \times C$ to $\alpha \times N \times d$, to compute the impact of anomaly factors on sensor space. Second, using two fully connected layer, we generate the query and key for each sensor at time, allowing us to compute the similarity between different sensors and obtain the similarity matrix $A_{S}\in R^{N \times N}$.
\begin{equation}
	A_{S} = \frac{Q_t^S{{(K}_t^S)}^\mathcal{T}}{\sqrt{\tilde{d}}}
\end{equation}
where $ Q_t^S=Exc\_T_{t::}W_q^S \in \mathbb{R}^{d \times \tilde{d}}$ represents the spatial query matrice obtained through the fully connected layer, $K_t^S=Exc\_T_{t::}W_k^S \in \mathbb{R}^{d \times \tilde{d}}$ indicates the spatial key matrice obtained through the fully connected laye.

Next, we construct a matrix $Mask$. When the number of hops between two nodes in the adjacency matrix $\mathbf{A}$ exceeds a certain threshold, the corresponding $Mask$ values are set to True. The positions in the similarity matrix that are True in the mask matrix are then set to negative infinity to eliminate interference from distant nodes, effectively capturing the spatial influence of anomaly factors. 

Finally, after applying the mask to the similarity matrix, we multiply it by the value matrix $V_t^S$ obtained from the fully connected layer to derive the temporal influence of external anomaly factors. The specific representation is as follows:
\begin{equation}
	\text{SSA}({Exc\_T}_{i::}, Mask)=FC(softmax(A_{T})V_t^S\ )
\end{equation}
where $V_t^S = Exc\_T_{t::}W_v^S \in \mathbb{R}^{N \times d}$ represents the time value matrix obtained from the fully connected layer. By applying the above operations to all sensor data, the output of the SSA component is obtained$ Exc_TS\in \mathbb{R}^{N \times \alpha \times d} $.
\begin{equation}
	Exc\_TS=\text{SSA}(Exc\_T, Mask)
\end{equation}
(4) Anomaly Detection Network

To measure the impact of different categories of external anomaly factors on traffic prediction, we designed an anomaly detection network. First, we predefine $M$ (where \(M\) is a hyperparameter in the deep classification network), which consists of $M$ learnable $d$-dimensional variables to store representations of different categories of anomaly factors. 

Next, we calculate the similarity between the input anomaly factors and the predefined $M$ categories of anomaly information, and then normalize the obtained similarity $\gamma_m^i$. This can be represented as:
\begin{equation}
	\gamma_m^i = \frac{e^{Exc\_TS[i] \cdot \omega_T[m]}}{\sum_{m'}^{M}{e^{Exc\_TS[i] \cdot \omega_T[m']}}} \quad \\
\end{equation}

Finally, we apply weighted processing to the normalized result and then feed it into a fully connected layer to obtain the enhanced external anomaly impact ${Exc\_inf}_i$, which can be represented as:
\begin{equation}
	{Exc\_inf}_i = \text{FC}(Exc_i)
\end{equation}
where $Exc_i = \sum_{m}^{M}{\gamma_m^i \ast \omega_m}$ indicates the weighted value of $i$-th anomaly category, $\omega_m$ refers to the vector representation of the $m$-th anomaly category.
\\
\\
\subsection{Data Embedding Module}
To enrich the information in the input sequence, we introduced various types of embedding data. First, we use the same fixed positional encoding based on sine and cosine functions as in the Transformer model to obtain the embedded data, denoted as $X_t \in \mathbb{R}^{\alpha \times d}$.
\begin{equation}
	\begin{cases}
		\text{PE}(\text{t}, 2i) = \sin\left(\text{t} / 10000^{(2i / d)}\right) \\
		\text{PE}(\text{t}, 2i + 1) = \cos\left(\text{t} / 10000^{(2i / d)}\right)
	\end{cases}
\end{equation}
where $ t $ represents the position in the time step, $ i $ is the index of the embedding dimension, and $ d $ is the total embedding dimension in the Transformer. 

In addition, traffic patterns exhibit significant periodicity. To capture this periodicity, we introduced daily embeddings $X_d \in \mathbb{R}^{\alpha \times d}$ and weekly embeddings $X_w \in \mathbb{R}^{\alpha \times d}$. Here,  we index each day from 1 to 1,440 and apply a linear transformation to obtain the mapping for each index, $X_d$. Similarly, we index each week from 1 to 7 and apply a linear transformation to derive the corresponding mapping, $X_w$. 

Moreover, to represent the graph structure in traffic data, we use the graph Laplacian eigenvectors to retain global information. We take the $ k $ smallest non-trivial eigenvectors from the Laplacian matrix and linearly project them to obtain $X_s \in \mathbb{R}^{N \times d}$. 

Finally, we add the obtained embeddings of each category to get the output $X_{emb}\in \mathbb{R}^{\alpha \times N \times d}$ of the data embedding module. This can be represented as:
\begin{equation}
	X_{emb}=X_{origin}+X_t+X_d+X_w+X_s
\end{equation}
where $X_{origin}$ is the high-dimensional embedding of the original data $X$.
\subsection{Multi-Scale Spatiotemporal Feature Fusion Module}
Previous methods often capture correlations between sensors by separating time and space, or are limited to short-term spatio-temporal correlations, resulting in an inability to fully capture the long-term correlations between different sensors. To address this issue, we designed a Multi-scale Temporal-Spatial Feature Fusion Module (MTSFFM) based on the Transformer architecture. This module comprises $L$ layers of Multi-scale Temporal-Spatial Feature Fusion Layer (MTSFFL), with each layer of MTSFFL containing three components: SSA, TSA, and Multi-scale Temporal-Spatial Feature Fusion component(MTSFF).

(1) MTSFF

To directly capture the interactions between different sensors over long durations, we constructed the MTSFF component to extract the correlations among different sensors in a long-term context. First, we define the input as tensor $H \in R^{\alpha \times N \times d}$, then we unfold it along the temporal dimension to obtain $H_{ts} \in R^{\alpha N \times d}$. Using three fully connected layers, we generate query, key, and value tensors corresponding to traffic sensors across different times and spatial locations.
\begin{equation}
		Q_{ts} = H_{ts}W_q^{ts}, 
		K_{ts} = H_{ts}W_k^{ts}, 
		V_{ts} = H_{ts}W_v^{ts} 
\end{equation}

Next, we calculate the influence matrix among all sensors across different time points $A_{st}\in R^{\alpha N \times \alpha N}$.
\begin{equation}
	A_{st}=\frac{Q_{ts}{{(K}_{ts})}^\mathcal{T}}{\sqrt{\tilde{d}}}
\end{equation}
Finally, since only a few interactions between nodes are necessary in practice, we expand the original tensor to a size of $(E, F)$ to reduce the influence of weakly correlated nodes. Specifically, the proposed formula for the masked spatio-temporal attention mechanism is as follows:
\begin{equation}
\small 	MTSFF(H_{ts},Mask)=FC(\text{softmax}(A_{st}\cdot Mask)V_{ts})
\end{equation}
(2) SSA And TSA components

Although the spatio-temporal hybrid self-attention component can capture the relationships between any sensors, when calculating sensor correlations, it only uses the sensor's own characteristics without considering the trend information of the sensors. To effectively capture the local trend between sensors, we introduced SSA and TSA components before the spatio-temporal hybrid self-attention component to extract local trend information.

First, we use SSA to aggregate local spatial features from the sensors.
\begin{equation}
	H_{s}^{i}= SSA(H^{i}, Mask)
\end{equation}
where $H^{i}$ represents the input vector of the $i$-th layer MTSFFL , and $H_{s}^{i}$ denotes the output vector of the SSA in the $i$-th layer. 

Next, we make use of TSA to aggregate the local temporal features from the sensors:
\begin{equation}
	H_{t}^{i}= TSA(Reshape(H_{s}^{i}))
\end{equation}
where $H_{t}^{i}$ indicates the spatio-temporal node representation after the local spatio-temporal information has been aggregated. 

Finally, the data containing local trend information is fed into the MTSFF component to obtain the output $H_{out}^{i}$ of the $i$-th layer, and $H_{out}^{i} \in \mathbb{R}^{\alpha N \times \tilde{d}}$ can be expressed as 
\begin{equation}
	H_{out}^{i}= MTSFF(Reshape(H_{s}^{i}), Mask) 
\end{equation}
(3) Skip Connection

To introduce richer multi-level features, we set up skip connections after each MTSFFL. The input to the first MTSFFL is $H^{1}$.
\begin{equation}
	H^{1}= X_{emb} + Exc_{inf}
\end{equation}
where $X_{emb}$ represents the output from the data embedding module, and $Exc_{inf}$ represents the output of the anomalous factors impacting  module. 

After processing through the three components, the output for the $i$-th layer is  $ $
\begin{equation}
	H^{i+1}= Reshape(H_{out}^{i})\in R^{\alpha \times N \times d}
\end{equation}

Finally, after reshaping the output of each layer, we aggregate them to obtain the final hidden state, denoted as $ \bar{X} $.
\begin{equation}
	\bar{X}= Reshape(H_{out}^{1}) + ... + Reshape(H_{out}^{L})
\end{equation}

To produce the final prediction results, we use two $1 \times 1$ convolutional neural networks to transform the temporal dimension and the feature dimension to the desired output size. The exact transformation is illustrated in the following formula:
\begin{equation}
	\bar{Y}= con2(con1(\bar{X}))\in R^{\beta \times N \times C}
\end{equation}
where $ \bar{Y} $ represents the predicted output, $ \beta $ denotes the number of time steps for the prediction, and $ C $ indicates the number of categories in the traffic state.
\section{EXPERIMENTAL EVALUATION}
\subsection{Environment}
The efficacy of our method was evaluated on two publicly available datasets, PEMS04 and PEMS08. The datasets were partitioned into training, validation, and test sets in a 6:2:2 ratio. The model was implemented with PyTorch 1.11.0+cu113 and executed on an NVIDIA RTX 4070Ti GPU. The identical parameters were employed for both datasets. As illustrated in Table \ref{PARAMETER}: In the anomaly detection network, the number of anomaly categories, denoted by $M$, is 64. Each category is represented by a learnable 64-dimensional vector. The hidden dimension in the self-attention mechanism is set to 16. The observation step size and prediction horizon are both set to 12. The number of spatio-temporal feature extraction layers is four. Adam optimizer is employed with a learning rate of 0.01, and the batch size is 64. Early stopping occurs if the validation error converges within 50 epochs or if the training process reaches 400 epochs. Root Mean Square Error (RMSE), Mean Absolute Error (MAE), and Mean Absolute Percentage Error (MAPE) are utilized as evaluation metrics.

\begin{table}[htbp]
	\caption{PARAMETER SETTINGS}
	\begin{center}
		\renewcommand{\arraystretch}{1.5}
		\begin{tabular}{|>{\centering\arraybackslash}p{5.0cm}|>{\centering\arraybackslash}p{2.0cm}|}
			\hline
			\textbf{Parameter} & \textbf{Value} \\
			\hline
			Learning Rate & 0.01 \\
			\hline
			Batch Size ($B$) & 16 \\
			\hline
			Epochs & 400 \\
			\hline
			Hidden States ($d$) & 64 \\
			\hline
			Hidden States ($\tilde{d}$) & 16 \\
			\hline
			Historical sequence step ($\alpha$) & 12 \\
			\hline
			Prediction Horizon ($\beta$) & 12 \\
			\hline
			The number of feature fusion layers ($L$) & 4 \\
			\hline
			The number of anomaly categories ($M$) & 64 \\
			\hline
			Early stopping occurs & 50   \\
			\hline
			The number of traffic states & 1   \\
			\hline	
		\end{tabular}
		\label{PARAMETER}
	\end{center}
\end{table}
\subsection{Quantitative Evaluation}
We chose a selection of traditional methods (such as HI), typical deep learning methods (e.g., GWNet\cite{GWNet}, DCRNN\cite{DCRNN}, AGCRN\cite{AGCRN}, STGCN\cite{STGCN}, GTS\cite{GTS}), and state-of-the-art methods (e.g., GMAN\cite{GMAN}, MTGNN\cite{MTGNN}, STID\cite{STID}, STNorm\cite{STNorm}, PDFormer\cite{PDFormer}, STAEformer\cite{STAE}). From Table \ref{evaluation_metrics}, we can observe that our proposed model outperforms the latest state-of-the-art methods across all metrics. STAEformer introduced a novel concept of spatio-temporal embedding, comprised of a set of learnable parameters. While it captures different representations for various nodes at different times, it does not explicitly extract relationships between these spatio-temporal nodes, nor does it account for the influence of external anomalies. As shown in the table, compared to STAEformer, our method reduces MAE by 3.57\%, RMSE by 5.81\%, and MAPE by 3.49\%. This indicates that our approach's ability to capture long-term global spatio-temporal features and model external anomalies effectively reduces errors in traffic prediction.
\begin{table}[htbp]
	\centering
	\caption{Evaluation Metrics on PEMS04 and PEMS08 Datasets}
	\label{evaluation_metrics}
	\resizebox{\columnwidth}{!}
	{ 
		\renewcommand{\arraystretch}{1.3}
		\begin{tabular}{|c|c|c|c|c|c|c|}
			\hline
			Method & \multicolumn{3}{c|}{PEMS04} & \multicolumn{3}{c|}{PEMS08} \\ \hline
			Metric & MAE       & RMSE       & MAPE       & MAE       & RMSE      & MAPE       \\ \hline
			HI     & 42.35     & 61.66      & 29.92\%    & 36.66     & 50.45     & 21.63\%    \\ \hline
			GWNet  & 18.53     & 29.92      & 12.89\%    & 14.40     & 23.39     & 9.21\%     \\ \hline
			DCRNN  & 19.63     & 31.26      & 13.59\%    & 15.22     & 24.17     & 10.21\%    \\ \hline
			AGCRN  & 19.38     & 31.25      & 13.40\%    & 15.32     & 24.41     & 10.03\%    \\ \hline
			STGCN  & 19.57     & 31.38      & 13.44\%    & 16.08     & 25.39     & 10.60\%    \\ \hline
			GTS    & 20.96     & 32.95      & 14.66\%    & 16.49     & 26.08     & 10.54\%    \\ \hline
			MTGNN  & 19.17     & 31.70      & 13.37\%    & 15.18     & 24.24     & 10.20\%    \\ \hline
			STNorm & 18.96     & 30.98      & 12.69\%    & 15.41     & 24.77     & 9.76\%     \\ \hline
			GMAN   & 19.14     & 31.60      & 13.19\%    & 15.31     & 24.92     & 10.13\%    \\ \hline
			PDFormer & 18.36    & 30.03      & 12.00\%    & 13.58     & 23.41     & 9.05\%     \\ \hline
			STID   & 18.38     & 29.95      & 12.04\%    & 14.21     & 23.28     & 9.27\%     \\ \hline
			STAEformer & 18.22  & 30.18      & 12.01\%    & 13.46     & 23.25     & 8.88\%     \\ \hline
			GSTF & \textbf{17.79} & \textbf{28.80} & \textbf{11.98\%} & \textbf{12.98} & \textbf{21.90} & \textbf{8.57\%} \\
			\hline
		\end{tabular}
	}
\end{table}
\subsection{Ablation Study}
\begin{figure}[ht]
	\centering
	\includegraphics[width=0.9\columnwidth]{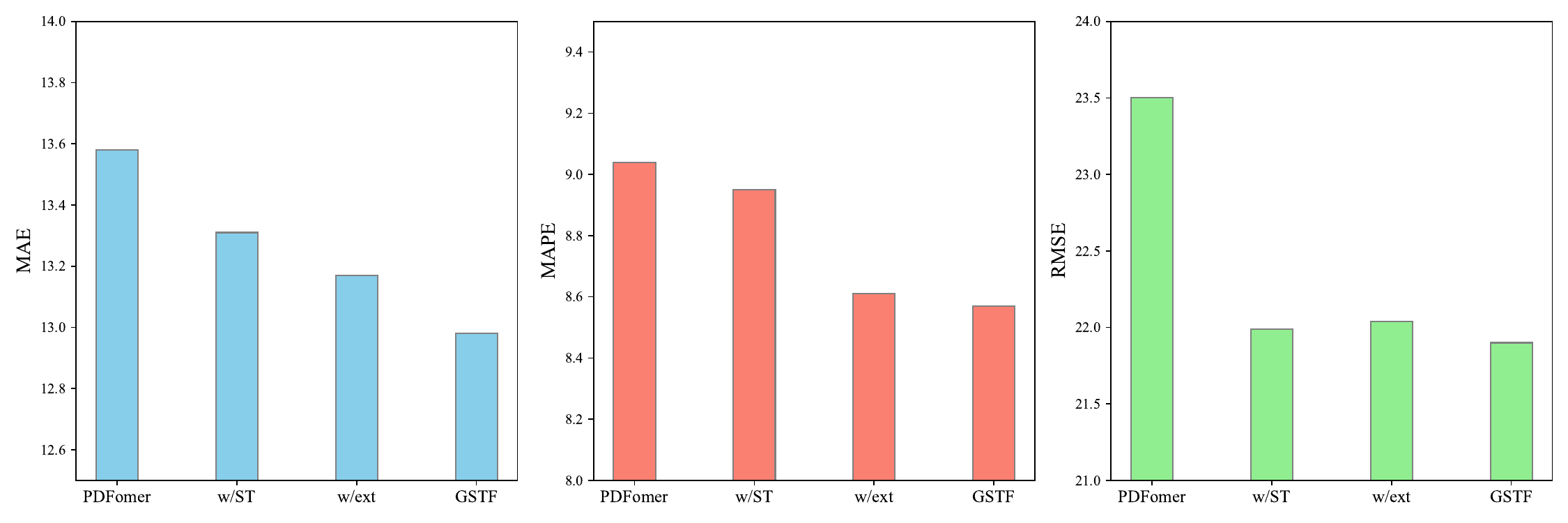}  
	\caption{Ablation Study on the PEMS08 Dataset.}
	\label{Ablation}
\end{figure}

To further examine the effectiveness of different modules in our method, we compared the GSTF module with the following variants: (1) PDFormer: This method is based on a transformer architecture, and we use it as a baseline model. (2) w/ST: This variant removes the multi-scale spatiotemporal fusion component from the multi-scale spatiotemporal feature fusion module. (3) w/ext: This variant removes the anomalous factor impact module.  Fig \ref{Ablation} shows a comparison of these variants on the PEMS08 dataset, we can see that our proposed multi-scale spatio-temporal feature fusion module significantly improves accuracy, which implies that the relationships among  different sensors over a long period have a significant impacts on prediction tasks. Additionally, it can be observed that our proposed anomaly factor module effectively reduces the RMSE value. Since the RMSE value is more sensitive to outliers compared to the other two metrics, this suggests that effectively modeling external anomaly factors can greatly reduce the impact of outliers on the model.
\section{Conclusion and Future Work}
In this paper, we have proposed a global spatiotemporal fusion-based traffic prediction algorithm. First, based on an anomaly classification network, we designed an anomalous factors impacting  module that effectively represented the spatio-temporal impact of external anomalies on traffic prediction. Additionally, we constructed a multi-scale spatio-temporal feature fusion model based on the Transformer architecture, which can capture all potential long-term and short-term correlations among different sensors in a broad traffic environment, enabling accurate traffic flow prediction. Experiments conducted on two real-world datasets validated the superiority of our approach. In the future work, we will further study how to improve the robustness of traffic prediction models and reduce model parameters and training time.
\bibliography{IEEEabrv.bib}
\end{document}